\documentclass[10pt]{article}
\usepackage[legalpaper, margin=1in]{geometry}
\usepackage{amsmath,amsfonts}
\usepackage{algorithmic}
\usepackage{algorithm}
\usepackage{array}
\usepackage[caption=false,font=normalsize,labelfont=sf,textfont=sf]{subfig}
\usepackage{textcomp}
\usepackage{stfloats}
\usepackage{url}
\usepackage{verbatim}
\usepackage{graphicx}
\usepackage{cite}
\usepackage{xcolor}
\usepackage{multirow}
\usepackage{wrapfig}

\begin{document}


\title{The Stanford Drone Dataset is More Complex than We Think: An Analysis of Key Characteristics}

\author{Josh Andle, Nicholas Soucy, Simon Socolow, and Salimeh Yasaei Sekeh}%




\date{}
\maketitle

\begin{abstract}
Several datasets exist which contain annotated information of individuals’ trajectories. Such datasets are vital for many real-world applications, including trajectory prediction and autonomous navigation. One prominent dataset currently in use is the Stanford Drone Dataset (SDD)~\cite{robicquet2016learning}. Despite its prominence, discussion surrounding the characteristics of this dataset is insufficient. We demonstrate how this insufficiency reduces the information available to users and can impact performance. Our contributions include the outlining of key characteristics in the SDD, employment of an information-theoretic measure and custom metric to clearly visualize those characteristics, the implementation of the PECNet \cite{mangalam2020not} and Y-Net \cite{mangalam2021goals} trajectory prediction models to demonstrate the outlined characteristics' impact on predictive performance, and lastly we provide a comparison between the SDD and Intersection Drone (inD) Dataset. Our analysis of the SDD’s key characteristics is important because without adequate information about available datasets a user’s ability to select the most suitable dataset for their methods, to reproduce one another’s results, and to interpret their own results are hindered. The observations we make through this analysis provide a readily accessible and interpretable source of information for those planning to use the SDD. Our intention is to increase the performance and reproducibility of methods applied to this dataset going forward, while also clearly detailing less obvious features of the dataset for new users.


\end{abstract}



\section{Introduction}
\label{sec:intro}

Several datasets are available which provide annotated trajectory data of individuals navigating one or more scenes~\cite{pellegrini2009you, lerner2007crowds, robicquet2016learning, bock2020ind, krajewski2018highd}. Such datasets can be used as tools for an array of problems in Computer Vision and Intelligent Vehicles, including object detection~\cite{yang2019effective}, object tracking~\cite{sadeghian2017tracking}, and trajectory prediction~\cite{lee2017desire, mangalam2021goals, zhao2020tnt}. A popular dataset that has been widely used for comparing recent benchmark methods is the Stanford Drone Dataset (SDD)~\cite{robicquet2016learning}. The frequent use of the SDD stems in part from its size, the fact that it contains both vehicles and pedestrians in crowded scenes, and its previous use by benchmark methods. While many papers use the SDD to evaluate their performance, often the clear justifications for selecting it over alternative datasets is lacking.

The scarcity of information detailing the SDD may make it harder for users to properly determine if it's the best fit for their research. This scarcity is largely due to the fact that the dataset's accompanying publication and documents do not provide users with a sufficiently comprehensive description of the dataset, nor do the papers which utilize it. 

One characteristic which is not addressed is how the videos within the SDD relate to one another in terms of time and location of recording. Another feature which plays a major role in determining the suitability of the SDD for a given model is the actual distribution and behavior of classes within the dataset compared to other, similar dataset options. Understanding differences in distribution and behavior is important when making an informed choice regarding which dataset is most suited to a particular application. Lastly, we discuss properties of the annotation data, including the impact of annotations labeled "lost" and split trajectories that effect tracking and prediction models of agents both directly and indirectly. For methods like trajectory prediction which rely on the annotated coordinates data, properly understanding these characteristics is necessary to ensure that the results are correct and easily reproducible. 

\begin{figure*}[t]
\centering
\hspace*{-0cm}\includegraphics[scale=0.9]{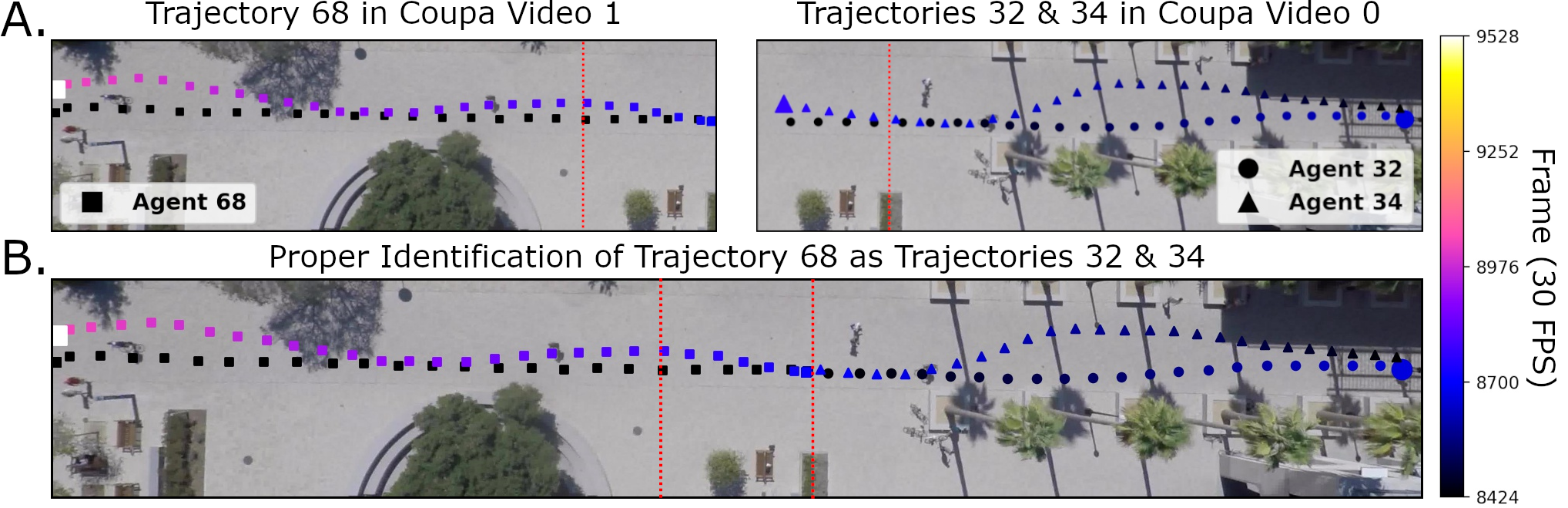}
\caption{A. Three trajectories are shown from the SDD's Coupa videos 1 (left) and 0 (right). Annotations show agent 68 exiting the right side of video 1 and returning shortly after. The intermediate frames are labeled as "lost", indicating that the agent is out of the video's bounds. B. Knowing that videos 0 and 1 are shot simultaneously and overlap in location, it becomes clear that agent 68 is actually a combination of agent 32 and 34's trajectories. Since agent 34 enters video 1 at the same location as agent 32 leaves, they are mistakenly annotated as the same individual in video 1 with their trajectories connected by "lost" annotations.}
\label{fig.Complexity}
\end{figure*}

We demonstrate each of these characteristics, and showcase how they may impact the accuracy of trajectory prediction applications. In order to compare them to a similar dataset we include the Intersection Drone (inD) dataset within the scope of our investigation. An example of the complexity that these characteristics add to the SDD can be seen in Fig.~\ref{fig.Complexity}. The characteristics of importance demonstrated in this example are the way in which scenes are oriented relative to one another, as well as the behavior of "lost" annotations. Together, knowledge of these characteristics allows the user to determine the erroneous nature of the annotations in the provided example. This illustrates how understanding the characteristics of the dataset discussed here can provide a more comprehensive understanding of the SDD, and with it an improved ability to interpret observations made about the dataset.

In order to visualize and validate the outlined key characteristics, we implement the pre-existing PECNet~\cite{mangalam2020not} and Y-Net~\cite{mangalam2021goals} trajectory prediction models, as well as a custom Adaptive Interaction Measure (AIM). AIM utilizes the information-theoretic measure of mutual information (MI)~\cite{TC}, which has been used in previous works involving tracking and trajectory prediction problems~\cite{xu2015misfm, zhang2017vehicle}. We utilize AIM to visualize various dataset characteristics, and apply PECNet and Y-Net to the SDD to demonstrate some of those characteristics' impact on performance when ignored or accounted for. We chose to use these models due to the functional clarity, accurate performance, and readily available code.

This paper's aim is to highlight information about the SDD which is not readily apparent or discussed in previous work, as well as to demonstrate the importance of that information. Since there are several applications which the SDD can be used for, an analysis of exactly how those characteristics impact each possible application is outside of the scope of this paper. Instead we aim to utilize only as many experiments and methods as are necessary to provide compelling evidence of these characteristics' importance. Additionally, the custom AIM measure is intended as a metric to visualize and evaluate these characteristics. Analyzing and rigorously investigating the AIM measure is out of our scope in this research. In summary, our contributions in this paper are as follows:
\begin{enumerate}
    \item {We describe the "lost"-labeled annotations and split trajectories within the SDD and their importance. We demonstrate how properly handling these occurrences during preprocessing impacts the resulting model performance.}
    \item {We provide a holistic view of how the videos within a given scene fit together. This includes cases where videos are recorded simultaneously, overlap in location, or both. We propose situations in which understanding how the videos relate to one another would impact the results obtained with the dataset.}
    \item {We demonstrate differences in class distribution and behavior between the SDD and inD dataset which should be taken into consideration when deciding when to use one or the other.}
\end{enumerate}

To the best of our knowledge, there is no research which provides sufficient information on how the authors processed the SDD data and how the characteristics we discuss in this work might have impacted their results and model evaluation.

\section{Related Works}
\label{sec:related-works}
Two datasets which have previously been used as benchmarks are ETH Zurich's BIWI Walking Pedestrian Dataset~\cite{pellegrini2009you} and UCY's Crowds by Example~\cite{lerner2007crowds}. These datasets are often used together as the ETH/UCY dataset. However, the small number of videos and lack of non-pedestrian individuals limit their utility, and differences in annotation formats between the two datasets could make using them together more troublesome for complex models. 

These datasets have since been largely replaced in favor of the Stanford Drone Dataset (SDD)~\cite{robicquet2016learning}. The SDD addresses several of the shortcomings of ETH/UCY, providing 60 videos split across 8 scenes and 6 classes of individuals, all with a single consistent annotation method. More recently the Intersection Drone (inD) dataset~\cite{bock2020ind} has aimed to improve upon the SDD, providing 33 videos split across 4 intersections and 4 classes of individuals. 

One prominent application for which the SDD has been used is trajectory prediction. Prominent examples of benchmark methods which use the SDD include Social-GAN~\cite{gupta2018social}, SoPhie~\cite{sadeghian2019sophie}, DESIRE~\cite{lee2017desire}, and CAR-Net ~\cite{sadeghian2018car}. PECNet~\cite{mangalam2020not} and Y-Net~\cite{mangalam2021goals} are two other models which use the SDD and report improved performance accuracy over previous baselines.

A notable resource which has helped mitigate inconsistent use of the SDD is TrajNet~\cite{sadeghian2018trajnet, kothari2021human}. This benchmark provides a more uniform method of preprocessing and performing trajectory prediction on various datasets, including SDD. 
Although TrajNet provides a uniform method of processing the SDD it doesn't provide a detailed analysis of the dataset which would allow users to properly understand it. Additionally, users of the dataset who wish to preprocess the data differently may not be able to rely on TrajNet. For these reasons, we provide a direct analysis of the dataset's characteristics. We focus on providing an intuitive visualization of a set of characteristics which users should be aware of when using the SDD and demonstrating their importance.



\section{Methods of Analysis}
In this section we introduce the methods used in our analysis of the SDD and inD dataset. One method we introduce is our information-theoretic metric to provide an interpretation of pairwise interactions between individuals, which we apply to visualize the characteristics of each dataset. Throughout the paper we use the notations below:
\begin{itemize}
    \item $X^I$: Trajectory coordinates for agent $I$.
    \item $X^I_{t_1:t_2}$: Trajectory coordinates for agent $I$ in time range $[t_1,t_2]$.
    \item $X^I_{t}$: Trajectory coordinates for agent $I$ at time point $t$.
    \item $\delta$: decay hyperparameter.
    \item $T$: Last time point of the interaction. 
    \item $T'$: A fixed number of time points after the beginning of the interaction.
\end{itemize}

\subsection{Datasets}

{\it Stanford Drone Dataset (SDD):} The SDD is a dataset providing birds-eye view drone recordings of 8 different scenes and 60 videos across a campus setting, with 6 annotated classes of individuals~\cite{robicquet2016learning}. These classes are pedestrians, skateboarders, bicyclists, carts, cars, and buses. The annotated data contains bounding-box coordinates in pixel values at 30 frames per second, as well as labels indicating if a given coordinate was occluded, "lost" out of the video's bounds, or automatically interpolated. \\

{\it Intersection Drone (inD) Dataset:} The inD dataset also provides birds-eye view drone recordings of 4 intersections across 33 recordings~\cite{bock2020ind}. Only the annotations are provided along with single-frame reference images, while the raw video footage isn't provided. The annotations contain 5 classes: pedestrians, bicyclists, cars, trucks, and buses, however the labels for trucks and buses are grouped as "truck\_buses" within the annotations. The annotations provide numerous additional parameters, including bounding box coordinates in meters at 25 frames per second, the necessary values to convert from meters to pixels, the speed limit, time and geographical location, as well as the individuals heading, velocity, and acceleration.

In order to make direct comparisons to the SDD we have converted all inD annotations to pixels using the provided conversion values prior to analysis. For the SDD we use the training/testing split outlined in \cite{sadeghian2018trajnet}. Table \ref{table:indsplit} outlines our split for the inD dataset, while later on in Section~\ref{sec.characteristics} Table \ref{table:classes} reports the class distribution for each dataset. For both datasets we preprocess the data to 2.5 fps and use the first 8 time-points of each trajectory for observation and the subsequent 12 time points for prediction, corresponding to 3.2 seconds and 4.8 seconds respectively.

\begin{table}[h]
\caption{ Dataset split used for the inD dataset.\\}
\centering
\scalebox{1.2}
{\vspace{1cm}
\begin{tabular}{@{}lcc@{}}
\cline{1-2}
{\bf Dataset} & {\bf Videos}\\
\cline{1-2}
Train & 0-4, 7-13, 18-25, 30\\ 
\cline{1-2}
Validation & 5, 14-15, 26-27, 31\\ 
\cline{1-2}
Test & 6, 16-17, 28-29, 32\\ 
\cline{1-2}
\end{tabular}}
\label{table:indsplit}
\end{table}

\begin{figure*}
\centering
\hspace*{0cm}\includegraphics[scale=0.38]{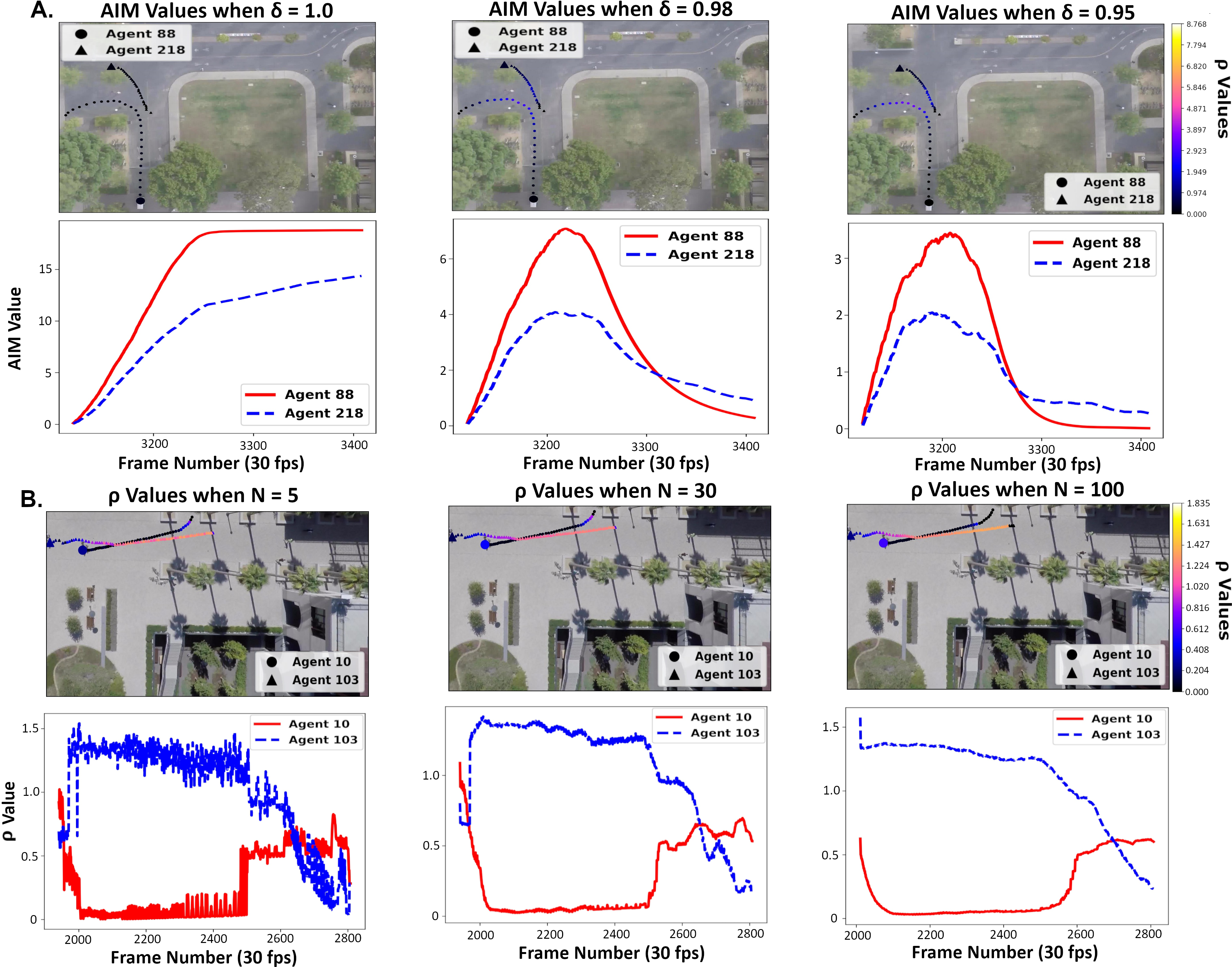}
\caption{A. We show the different values of AIM obtained for a single interaction when $\delta=1.0, 0.98,$ and $0.95$. When $\delta=1.0$, AIM doesn't decay and the summation is monotonically increasing. Introducing $2\%$ decay with $\delta=0.98$ results in a constant decay of the value of AIM. The overall value of AIM still increases when newly added time points outweigh the decrease caused by decay. Further decreasing to $\delta=0.95$ increases the rate of decay and observed AIM values, but doesn't substantially change behavior of AIM compared to $\delta=0.98$.} B. When $N=5$, $\rho$ varies erratically since $\frac{1}{5}$ of the input values are changed at each new time point. For $N=30$, only $\frac{1}{30}$ of the inputs change at each time point, which reduces the rapid changes seen when $N=5$. When $N=100$, the smoothing fails to capture the  brief changes that occur in the pedestrians' trajectories after frame 2400.
\label{fig.delta_n}
\end{figure*}

\subsection{AIM Definition}
In our {\it Adaptive Interaction Measure} (AIM) we incorporate a physics-based weight function, $\rho$, as a scalar value to weight mutual information (MI) based interactions~\cite{stuhl2022weighted, suhov2016basic, oselio2019time}. By summing this scaled information over the duration of a given trajectory pair, we get the cumulative measure, AIM, which describes the overall expected impact that one trajectory has on the other's navigation. The larger values of AIM suggest more significant interactions between agents. Our method is well-suited for this visualization task due in part to its transparent and intuitive incorporation of physical parameters in calculating $\rho$ and subsequently AIM.

Suppose $X^I$ and $X^J$ are the 2-dimensional trajectory coordinates for agents $I$ and $J$ for each time point in their interaction. The interaction between agents $I$ and $J$ is defined to contain all frames in which both individuals are in the scene together. The AIM for an interaction between two agents $I$ and $J$ over the frames $t=T',T'+1,\ldots, T$, denoted by $ AIM(X^I_{1:T};X^J_{1:T})$ is defined as follows

\begin{equation}
 \sum\limits_{t=T'}^{T}\delta^{T-t} \rho(X^I_{t-N:t},X^J_{t-N:t}) \; MI(X^I_{1:t};X^J_{1:t}),\label{AIM}
\end{equation}


We calculate MI and $\rho$ at each time point between $T'$ and $T$ and sum their product over all time points in the interaction. We provide the definitions of MI and $\rho$ in \ref{def:MI} and \ref{rho} respectively. At each step in the summation a constant decay term, $\delta$, is applied to the terms of each previous product. The AIM measure (\ref{AIM}) shows the result of expanding this function out, in which the decay term has a stacking effect on time points the further they are into the past of the trajectory. 

When $\delta=1.0$, no decay is applied to the past time points in the calculation of AIM, making it possible for high values of AIM to reflect interactions that are no longer relevant to present navigation decisions. When $\delta<1$, the value of AIM decreases at each time point. This allows more recent interactions to have a larger impact on the overall value of AIM, and in turn how an individual is expected to allocate their attention among neighboring individuals. 

To calculate $\rho$ at a given time point $t$, we consider the previous $N$ time points. As $N$ increases, the relative impact of replacing one frame at each new time point lessens, which has a smoothing effect on the calculated value of $\rho$. When $N$ is too large this smoothing removes changes of $\rho$ which reflect abrupt, temporary behaviors such as sudden stopping or breaking of a pedestrian or car. We demonstrate the effects of changing $\delta$ and $N$ in Fig. \ref{fig.delta_n}. In order to ensure that there is sufficient past information to properly calculate MI and $\rho$ we set aside the first $T'$ frames of interaction as a buffer and begin the summation at $t=T'$.


\begin{figure*}[t]
\centering
\includegraphics[scale=0.65]{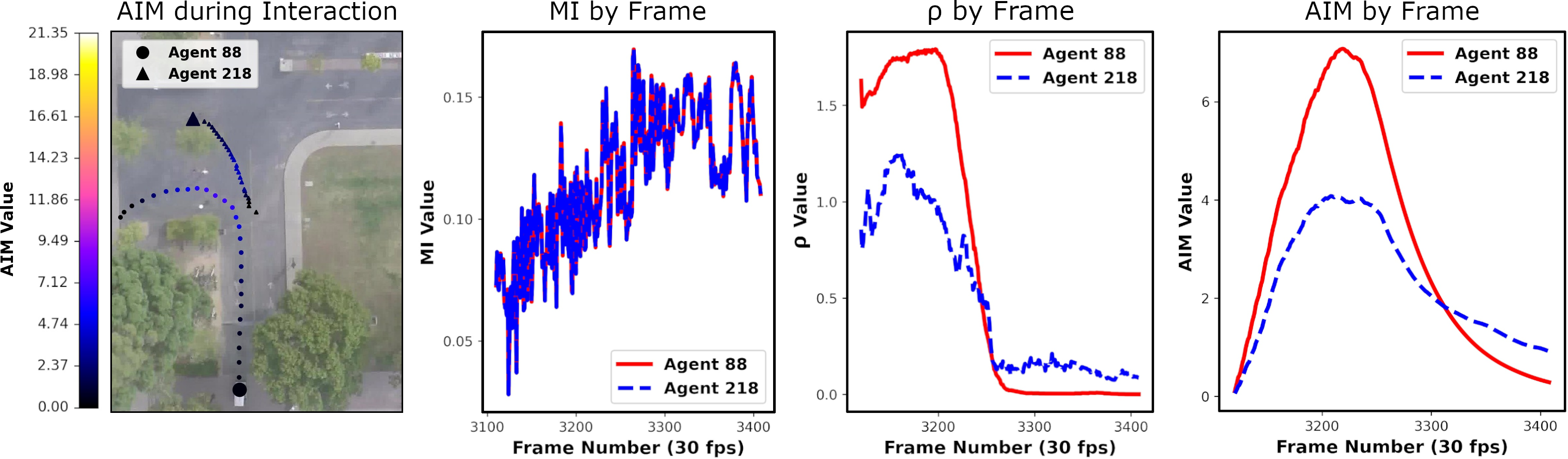}
\caption{For this example, we track MI, $\rho$, and AIM values for each frame in the trajectory pair. The car (agent 88) shows a higher initial $\rho$ value than the pedestrian, which is reflected by the fact that its AIM increases faster than the pedestrian's. Once the car has passed by the pedestrian, both $\rho$ values decrease and as a result AIM begins to decrease. This demonstrates that the expected attention is temporary in the absence of further relevant interactions.}
\label{fig.AIM}
\end{figure*}

\noindent{\bf Mutual Information:}
Let $P_{XY}$ be a probability measure on the  Euclidean space $\mathcal{X}\times \mathcal{Y}$. 
Here, $P_X$ and $P_Y$ define the marginal probability measures. The MI reflects the information that each of a pair of variables gives about the other, and is defined as
\begin{equation}\label{def:MI}
  MI(X;Y) = \mathop{\mathbb{E}}_{P_X P_Y}\left[  g\left(\frac{dP_{XY}}{dP_X P_Y}\right)\right], \;\;\;\;\; g(t)=\frac{(t-1)^2}{2(t+1)},
\end{equation}
where $\frac{dP_{XY}}{dP_X P_Y}$ is the Radon-Nikodym derivative. In this case, the variables are the coordinates along the trajectories of two individuals. High values of MI typically occur when agents are stationary or have been moving in a constant manner. MI does not noticeably change when the individuals are near or far apart, or based on the directions the individuals are moving in. To account for this we include our physics-based weighting measure $\rho$ when deriving AIM.

Examples of MI estimators include Kraskov Stogbauer Grassberger (KSG)~\cite{kraskov2004estimating}, Kernel Density Estimation (KDE)~\cite{moon1995estimation}, Nearest Neighbor Ratio (NNR)~\cite{noshad2017direct}, and Minimal Spanning Tree (MST)~\cite{yasaei2019geometric}. The MI estimation process is computationally intensive, e.g. the complexity of the KDE is $O(n^2)$, while the KSG takes $O(k n \log(n))$ ($k$ is a parameter). In this paper, we use a hash-based MI estimator called the ensemble dependency graph estimator (EDGE)~\cite{noshad2018scalable} due to its linear complexity and optimal mean squared error convergence rate.

\noindent{\bf Weight Function $\rho$:}
We utilize the handcrafted function $\rho$ in (\ref{AIM}) to provide a contextual interpretation of MI. We selected potential parameters of this function with the goal of reflecting the likelihood and severity of potential collisions between individuals. This is intended to approximate how attention is allocated by a navigating individual (e.g. avoiding oncoming cars, letting others pass on the sidewalk, and stopping at a crosswalk when people are crossing).
For the function's parameters, we use the {\it velocity} ($V$), {\it distance} ($D$), and the relative {\it heading} ($H$) of the two individuals. $V$ is the sum of both agents' velocities averaged over $N$ frames, $D$ is the average distance between the two agents, and $H$ is asymmetric and is calculated as the average angle between an individual's current heading and the other individual's current position. These parameters are calculated as follows:
\begin{align}
V_t  &= \frac{1}{N}\sum\limits_{n=t-N}^{t}\Big(\sqrt{\|X^I_n-X^I_{n-1}\|^2} +\sqrt{\|X^J_n-X^J_{n-1}\|^2}\Big)\label{V},\\
D_t  &= \frac{1}{N}\sum_{n=t-N}^{t}\sqrt{\|X^I_n - X^J_n\|^2},\\
H_t &= \frac{1}{N}\sum\limits_{n=t-N}^{t}\Big|arctan\left(\frac{\Big\|(X^J_{n-1} - X^I_{n-1}) \times (X^I_{n} - X^I_{n-1})\Big\|}{\Big<(X^J_{n-1} - X^I_{n-1}), (X^I_{n} - X^I_{n-1})\Big>}\right)\Big|,
\end{align}


where $\|.\|$ is Euclidean distance, the operation $\times$ is the cross product, and $<.\;,.>$ is the dot product. Prior to calculating $\rho$, $V_t$, $D_t$, and $H_t$ are normalized denoted by $V_{t^*}$, $D_{t^*}$, and $H_{t^*}$ respectively such that  $V_{t^*},\; D_{t^*}\in(0,1)$ while $H_{t^*}\in(-1,1)$.
The scale function which was used for $\rho$ is 
\begin{equation}
    \rho (X^I_{t-N:t}, X^J_{t-N:t}) := (\alpha + V_{t^*}) \cdot (D_{t^*}(1 + H_{t^*})).
    \label{rho}
\end{equation}

The constant $\alpha$ in (\ref{rho}) ensures that when $V=0$ then $\rho$ can still have a non-zero value based on $D_{t^*}$ and $H_{t^*}$; and when $V$ increases, so does $\rho$. By using $(D_{t^*}(1 + H_{t^*}))$, when one individual is moving toward the other's position $\rho$ increases. This reflects an increased likelihood of collision as the individuals move nearer to each other but also gives more attention to individual $J$ when they are in front of agent $I$, reflecting that agent $J$ would be readily visible to agent $I$ compared to neighbors outside of their field of view (assuming individuals look in the direction they're moving). 

\begin{figure}
\centering
\includegraphics[scale=0.39]{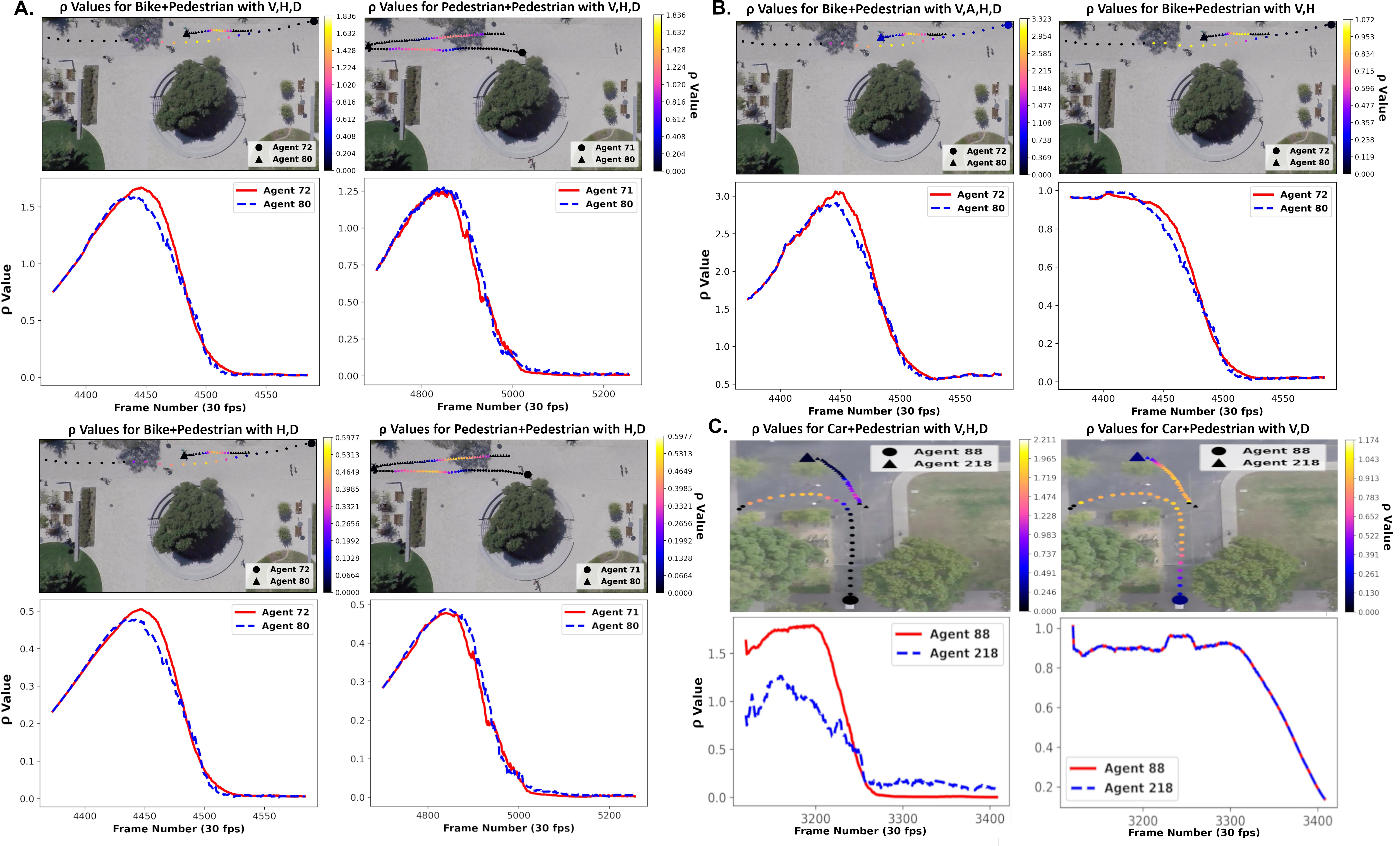}
\caption{A. Using {\it velocity} $V$ to calculate $\rho$ results in higher values for the interaction with agent 72, as they are moving faster than agent 71. Removing $V$ produces nearly the same values for both interactions, which is a less intuitive result. B. Introducing {\it acceleration} ($A$) does not significantly affect the behavior of $\rho$, while removing $D$ provides the same $\rho$ values regardless of distance, which is counter-intuitive for an attention measure. C. Removing {\it heading} $H$ from $\rho$ results in agents paying equal attention when either in front of or behind one another.}
\label{Ablative}
\end{figure}

In Fig. \ref{Ablative}, we demonstrate the impact of removing different parameters from the $\rho$ function. To do this, we compare two similar interactions between agents 80 (pedestrian), 71 (pedestrian), and 72 (bicyclist) as well as a third interaction between agents 88 (car) and 218 (pedestrian). We show that removing the velocity, distance, or heading parameters negatively impacts the intuitive interpretability of values of $\rho$ calculated for the interactions.


\subsection{Experimental Models}
For our experiments we use both the PECNet \cite{mangalam2020not} and Y-Net ~\cite{mangalam2021goals} models. PECNet is a trajectory prediction model which implements a variational auto-encoder and social pooling to predict trajectory endpoints. PECNet encodes the observed portions of trajectories and their ground-truth endpoints. This encoded information is used to predict end-points for each trajectory. The $K$ closest predictions to the goal are then used along with the observed portion of the trajectory to estimate the future coordinates of the trajectory. We use the author's provided code. Preprocessing was done directly with the SDD's annotations, using the PECNet model. Y-Net instead utilizes a set of sub-networks similar to U-net \cite{ronneberger2015u} to handle both the uncertainty in an agent's goal, as well as the uncertainty in how they reach a given goal. Encoding is performed for the past trajectory data and scene segmentation information, prior to predicting the future trajectory.

\subsection{AIM Implementation}
\label{sec.AIM.Interpret}

In this section we outline our hyperparameter settings for AIM and demonstrate how it can be applied to interaction data and interpreted. \\

{\it AIM Function:} In (\ref{AIM}), we set hyperparameter $\delta = 0.98$, resulting in a decay of $2\%$ at each time point.\\

{\it $\rho$ Function:} In our $\rho$ function (\ref{rho}) we set $\alpha=0.3$. We set $N=30$ for SDD and $N=25$ for inD. These values were selected based on the two dataset's differing frame rates, such that $\rho$ covers the past one second of trajectory data for each dataset. Each normalizing function for the parameters of $\rho$ was tuned with respect to the values of their parameter within the SDD.\\

We use MI, $\rho$, and AIM to visualize the characteristics of the SDD and inD dataset. Through the rest of the paper we present these measures as heatmaps overlaying the corresponding trajectories. The color of a given individual's trajectory indicates their value of the specified measure at that time point. The associated color bar reflects the corresponding values of the given measure among all individuals in a given recording. Accompanying these heatmaps are graphs tracking the values of the measure over the course of the trajectory for both interacting individuals. Fig.~\ref{fig.AIM} demonstrates this format while giving an example of how MI, $\rho$, and AIM vary over the course of a pair of trajectories.
The primary role of these measures within this paper is to provide readily interpretable methods of visualizing the characteristics of the SDD and inD datasets, rather than as tools for prediction, tracking or detection.


\section{Key Characteristics}
\label{sec.characteristics}

\subsection{Annotations}

\begin{figure*}
\centering
\hspace*{-0cm}\includegraphics[scale=0.415]{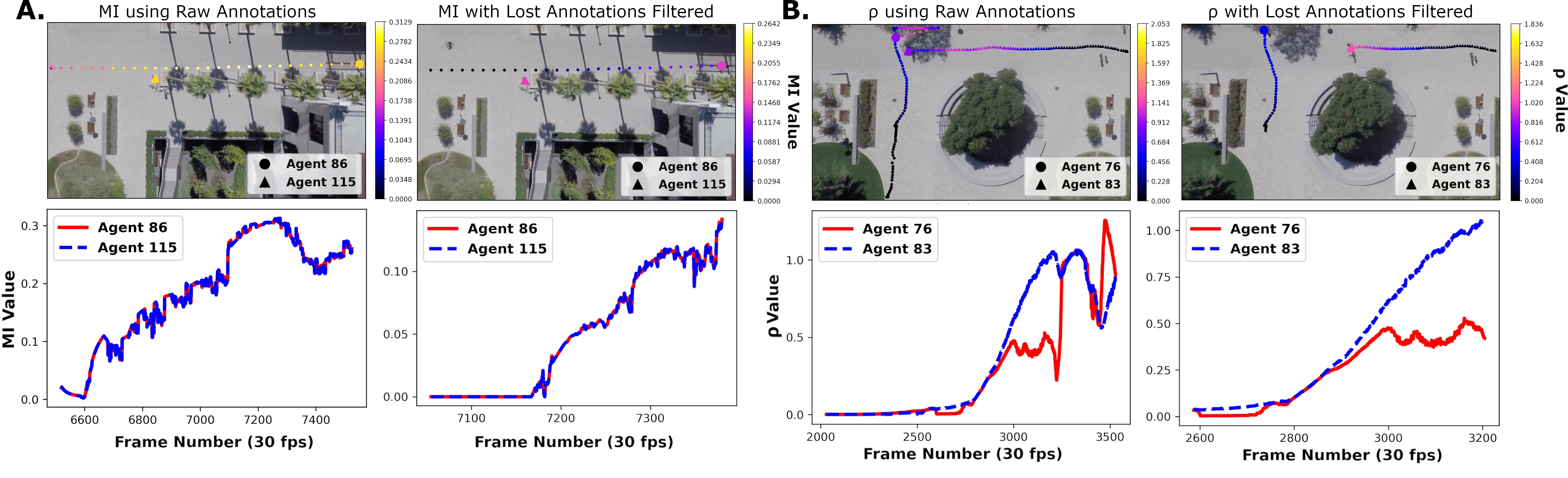}
\caption{A. The graphs reflect that nearly half of the trajectory's frames are removed when filtering "lost" annotations. As MI tends to increase during stationary trajectories, the additional frames inflate its value. B. Here the lost annotations wrongly suggest that the individual who has left the scene instead reenters and begins converging with the other individual. A predictive model may incorrectly predict that agent 83 will move to avoid a collision, when in fact there's no one else around them.}
\label{fig.LostAnnotations}
\end{figure*}

The SDD contains a label for frames that are "lost" out of the bounds of the video. This description alone doesn't give an adequate description of these annotations. When plotted, the actual behavior of these points becomes more apparent. Lost coordinates often occur either at the start or end of a trajectory, and less often in the middle for those agents that leave and reenter a scene. These annotations either remain stationary at the point where the individual will enter or leave the scene, or alternatively move linearly back into the scene. Fig.~\ref{fig.LostAnnotations} uses MI and $\rho$ to visualize some of the behavior and effects of these lost coordinates.

We report the frequency of trajectories containing these annotations in Table \ref{table:lost}. Additionally, we distinguish between how frequently they occur at the start of the trajectory, the end, or in the middle. This is important because simple approaches of filtering may properly remove the annotations at the start and end of the trajectory, but may have different effects on the middle occurrences. Some possible effects may be to leave a gap in the trajectory data with no coordinate data for those frames, or to split the trajectory into multiple new trajectories before and after the lost annotations occurred. For this reason it is important for users of the dataset to describe how they process the data they use. Omitting this information leads to difficulties when reproducing other researchers' work without explicit instructions on how they addressed such decisions. For this work we filter out the lost annotations, and then if this splits the trajectory we keep only the first portion. 

\begin{table}
\caption{Total number of trajectories and frequency of lost annotations in the SDD.}
\vspace{0.1cm}
\centering
\scalebox{.85}
{\begin{tabular}{@{}lccccc@{}}
\cline{1-5}
{\bf Scene} & {\bf Number of Trajectories} & {\bf Start(\%)} & {\bf Middle(\%)} & {\bf End(\%)}\\
\cline{1-5}
Bookstore & 1645 & 78.1 & 7.4 & 65.2\\ 
Coupa & 425 & 74.4 & 5.6 & 69.2 \\ 
DeathCircle & 2830 & 69.3 & 8.1 & 53.8 \\ 
Gates & 1249 & 68.9 & 10.6 & 54.8 \\ 
Hyang & 1980 & 68.7 & 9.9 & 56.2 \\ 
Little & 656 & 90.9 & 5.8 & 84.0 \\ 
Nexus & 1456 & 68.1 & 7.6 & 46.8 \\ 
Quad & 59 & 25.4 & 13.6 & 16.9 \\
\cline{1-5}
\end{tabular}}
\label{table:lost}
\end{table}

\begin{figure}
\centering
\hspace*{0cm}\includegraphics[scale=1]{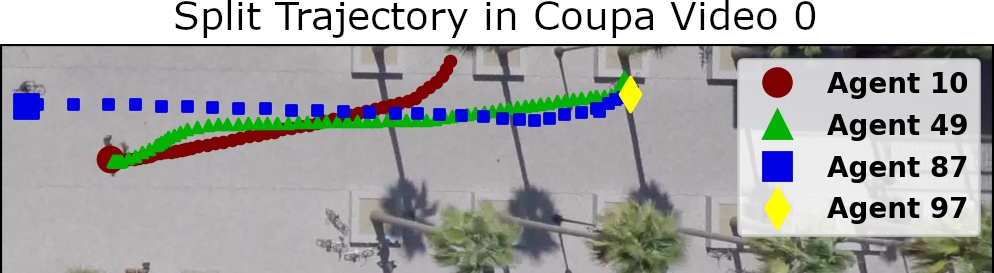}
\caption{In this example four unique track IDs from video 0 in the Coupa scene of the SDD are shown. The provided video and annotations confirm that these four tracks belong to a single individual.}
\label{fig.SplitTrajectory}
\end{figure}

This characteristic is of particular importance for researchers using the SDD for trajectory prediction, as many benchmark methods take $\sim{3.2}$ seconds of each trajectory as their observed portion of the trajectory and then attempt to predict the next $\sim{4.8}$ seconds~\cite{sadeghian2019sophie,lee2017desire,mangalam2020not}. If the lost coordinates are left included at the start of the trajectories, then this could lead to a large bias towards observing and predicting stationary trajectories. 

~\cite{lee2017desire, ridel2020scene} are the only works in which the lost coordinates are briefly mentioned, however the authors do not describe them. To demonstrate the importance of properly filtering these coordinates, we show how removing them can improve a trained trajectory prediction model's ability to generalize to a new dataset in Table \ref{pecnet}. This table reports the testing accuracy on the inD dataset using models trained either on the inD dataset, or on the SDD dataset with or without the lost annotations filtered out.


These results show that when training on the SDD and testing on the inD dataset, training on data which has the lost annotations filtered out (SDD w/o lost) improves both of the resulting performance measures (we use the standard trajectory prediction measures of Average Displacement Error (ADE) and Final Displacement Error (FDE) ~\cite{sadeghian2019sophie,kothari2021human,mangalam2020not}) compared to including them (SDD w/ lost). We show that this remains true when training/testing on all classes, pedestrians, bicycles, or cars, for both the PECNet and Y-Net models.

In addition to lost annotations, another behavior of the trajectory data in the SDD is that multiple trajectory IDs may correspond to a single individual. We demonstrate a prominent example of this behavior in Fig.~\ref{fig.SplitTrajectory}. In these cases the individuals full trajectory is split into multiple partial trajectories, each of which is given a different ID. Unlike with lost annotations, there is no clear indicator of when this occurs, and confirming it requires manual cross-checking of each partial trajectory's annotation.

\begin{table}
\centering
\caption{Testing accuracy on the inD dataset under different training conditions}
\vspace{0.1cm}
\scalebox{0.7}{
\begin{tabular}{cccccccccc} 
    \cline{1-10}
    Model & Dataset & \multicolumn{2}{c}{\bf{All}} & 
    \multicolumn{2}{c}{\bf{Pedestrians}} & \multicolumn{2}{c}{\bf{Bicycles}} & \multicolumn{2}{c}{\bf{Cars}} \\
    \cline{1-10}
    \hspace{0.01cm}
    & & ADE & FDE & ADE & FDE & ADE & FDE & ADE & FDE \\ 
    \cline{1-10}
    \hspace{0.01cm}
    \multirow{3}{*}{PECNet} & SDD w/ Lost & 21.82 & 35.26 & 7.77 & 13.95 & 26.87 & 44.11 & 83.25 & 149.32 \\
     & SDD w/o Lost & 20.18 & 35.08 & 7.16 & 12.46 & 21.8 & 33.11 & 65.38 & 112.33 \\
     & inD & 12.32 & 20.98 & 6.74 & 12.54 & 14.02 & 24.28 & 19.35 & 35.25 \\
    \cline{1-10}
    \multirow{3}{*}{Y-Net} & SDD w/ Lost & 14.22 & 22.61 & 8.76 & 13.51 & 20.80 & 29.00 & 72.10 & 64.55 \\
     & SDD w/o Lost & 13.84 & 22.28 & 7.49 & 12.37 & 19.61 & 26.80 & 82.75 & 41.96 \\
     & inD & 6.16 & 10.23 & 3.48 & 5.67 & 7.58 & 12.26 & 7.36 & 12.23 \\
    \cline{1-10}    
\end{tabular}}
\label{pecnet}
\end{table}

\subsection{Data Diversity and Scene Feature Adherence}
Previous works have commented on the SDD's class~\cite{krajewski2018highd} and scene~\cite{krajewski2020round} diversity. Vehicles make up a significant minority, and many of them are parked. In comparison, the inD dataset contains a larger percentage of moving cars but fewer pedestrians and classes.This suggests that the SDD may be more suited for applications that focus on pedestrians~\cite{dendorfer2020goal, mangalam2021goals}  while the inD dataset is suited for models intended for use in environments with both pedestrians and cars~\cite{cheng2021exploring} This idea is further supported by differences in scene diversity and navigational behavior. 

We demonstrate the percent distribution of classes in each dataset in Table \ref{table:classes}. The values for the SDD are provided by the datasets authors, while the values for inD were determined by checking the number of occurrences of each class in the metadata files and summing across all videos in a given intersection. These values show the prominence of foot traffic in the SDD, and of vehicles in the inD dataset.

\begin{table}
\caption{Class distribution in the SDD (Top) and the unique intersections of the inD dataset (Bottom).}
\centering
\scalebox{.8}
{\vspace{0.2cm}
\begin{tabular}{@{}lccccccc@{}}
\cline{1-8}
{\bf Dataset} & {\bf Scene} & {\bf Pedestrians} & {\bf Bicyclists} & {\bf Cars} & {\bf Buses} & {\bf Skaters} & {\bf Carts} \\
\cline{1-8}
\multirow{ 8}{*}{SDD} & Bookstore & 63.9 & 32.9 & 0.83 & 0.37 & 1.63 & 0.34\\ 
 & Coupa & 80.6 & 18.9 & 0.17 & 0 & 0.17 & 0.17\\ 
 & DeathCircle & 33.1 & 56.3 & 4.71 & 0.42 & 2.33 & 3.1\\ 
 & Gates & 43.3 & 51.9 & 1.08 & 0.78 & 2.55 & 0.29\\ 
 & Hyang & 70.0 & 27.7 & 0.5 & 0.09 & 1.29 & 0.43\\ 
 & Little & 42.5 & 56.0 & 0.17 & 0.67 & 0.67 & 0\\ 
 & Nexus & 64.0 & 4.22 & 29.5 & 1.25 & 0.6 & 0.4\\
 & Quad & 87.5 & 12.5 & 0 & 0 & 0 & 0\\    
\cline{1-8}
\multirow{ 4}{*}{inD} & 0-6 & 6.95 & 3.66 & 79.9 & 9.5 & 0 & 0 \\ 
 & 7-17 & 21.4 & 11.6 & 65.4 & 1.58 & 0 & 0 \\ 
 & 18-29 & 33.7 & 27.3 & 38.8 & 0.26 & 0 & 0\\ 
 & 30-32 & 3.44 & 3.05 & 88.9 & 4.61 & 0 & 0\\ 
\cline{1-8}
\end{tabular}}
\label{table:classes}
\end{table}

\begin{figure*}
\centering
\hspace*{-0cm}\includegraphics[scale=0.25]{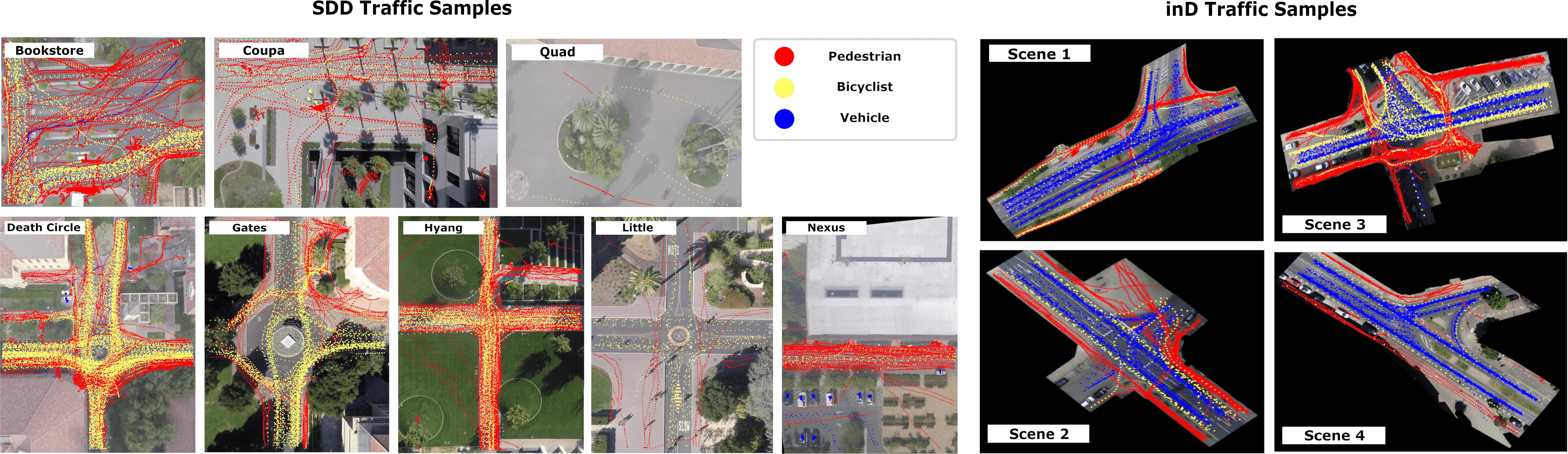}
\caption{The trajectories from a single video in each scene within the SDD and inD dataset are plotted. The behavior of foot traffic varies from scene to scene within the SDD. The pedestrians and bicyclists in the SDD often enter or cross the road, and vehicles are infrequent. In the inD dataset there are numerous vehicles, so bicyclists primarily remain on the shoulders of the road and pedestrians more consistently use crosswalks to cross.}
\label{fig.Traffic}
\end{figure*}

While the SDD has multiple scenes, all of these scenes are located in a campus setting where the predominance of foot traffic impacts navigation behavior. This behavior can be seen in Fig.~\ref{fig.Traffic}, which shows the frequency at which individuals in the SDD walk or bike through the streets rather than sidewalks or shoulders of the road. This indicates that certain scene features such as sidewalks and roads have a less predictable impact on navigation than when vehicles are more prevalent (as in the inD dataset). As such, the SDD may not be well suited to methods which learn to rely on such features.

In contrast to the SDD, the inD dataset has scenes located at four intersections in public roads. As with the SDD, there are some behaviors typical of intersections which are readily apparent in this dataset. For instance, cars are much more prone to stopping and turning than they would be in the absence of stop lights or intersections. Similarly, compared to areas such as the campus setting of SDD, cars are more prominent in the dataset, and as a result foot traffic has to adhere more closely to certain scene features than in SDD, as in Fig.~\ref{fig.Traffic}. For this reason, it's possible that the inD dataset is 
less suited for models which do not account for scene features, as these features add a constraint to navigational behavior.

An impact of the different behaviors and class distributions in both datasets can be seen in Table \ref{pecnet}. PECNet performed similarly well for pedestrians in the inD dataset whether it was trained on the SDD or the inD dataset, however the performance for cars is significantly worse when trained on the SDD, indicating that the cars in the SDD are insufficient for predicting car behavior in the inD dataset. This is possibly because many of the cars in the SDD are parked, or their driving is otherwise different due to the presence of more pedestrians in the road and lack of intersections with stoplights. We observe similar results with Y-Net where the gap in performance between pedestrians when trained on the SDD or inD dataset is much smaller than for cars.

\subsection{Scene Layout}

\begin{figure}
\centering
\includegraphics[scale=0.25]{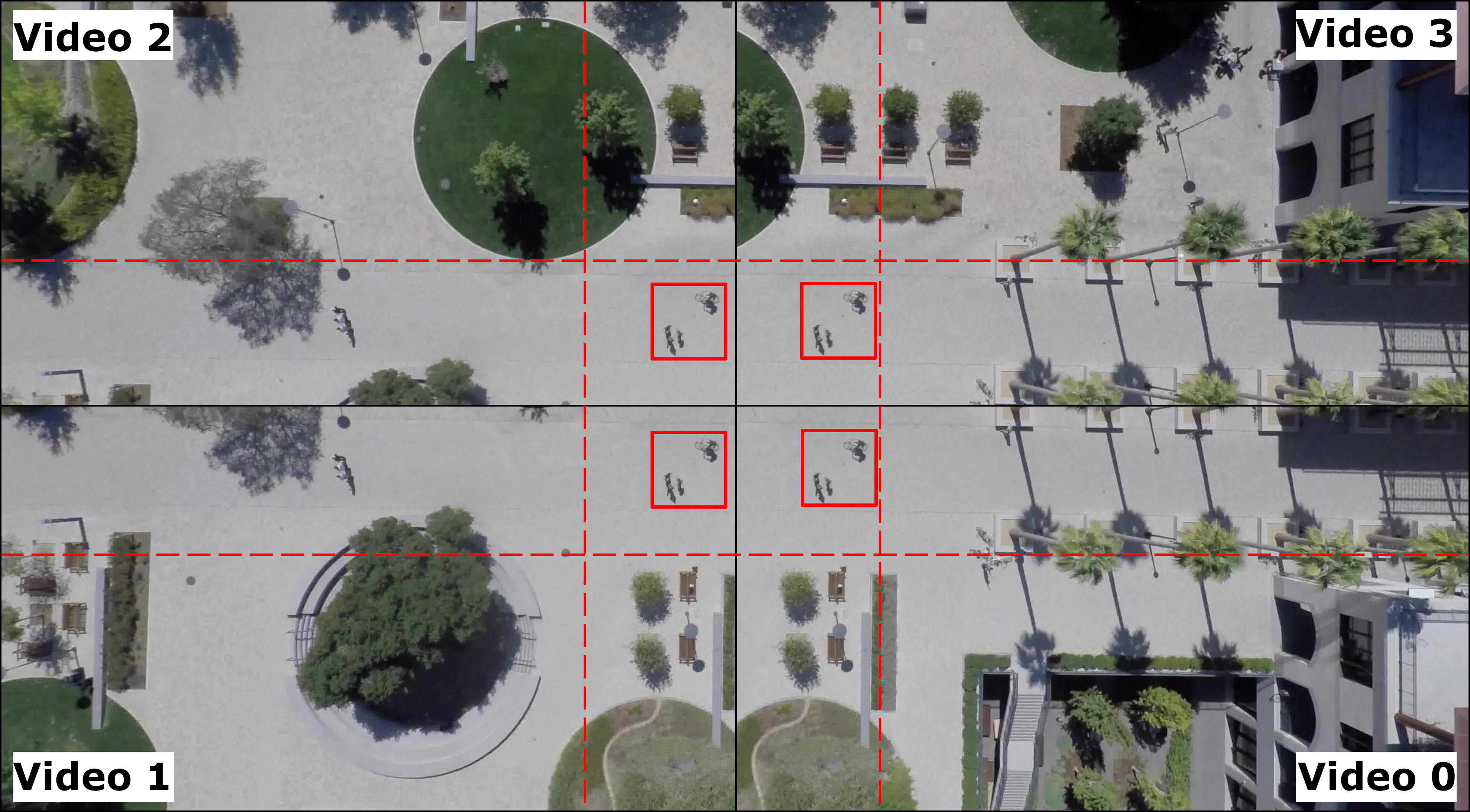}
\vspace{4mm}
\caption{Here the same frame is taken from each video in the Coupa scene. The overlap in location is reflected by dashed lines, and the simultaneous recording time can be seen from the four pedestrians who are in the same locations in each video.}
\label{fig.CoupaOverlap}
\end{figure}

For a given scene in the SDD, the region covered by each video may overlap completely, partially, or not at all. Similarly, the time of recording may be the same or different. An example in which both the location and time overlap is shown in Fig.~\ref{fig.CoupaOverlap}.

\begin{figure*}
\centering
\hspace*{0cm}\includegraphics[scale=0.415]{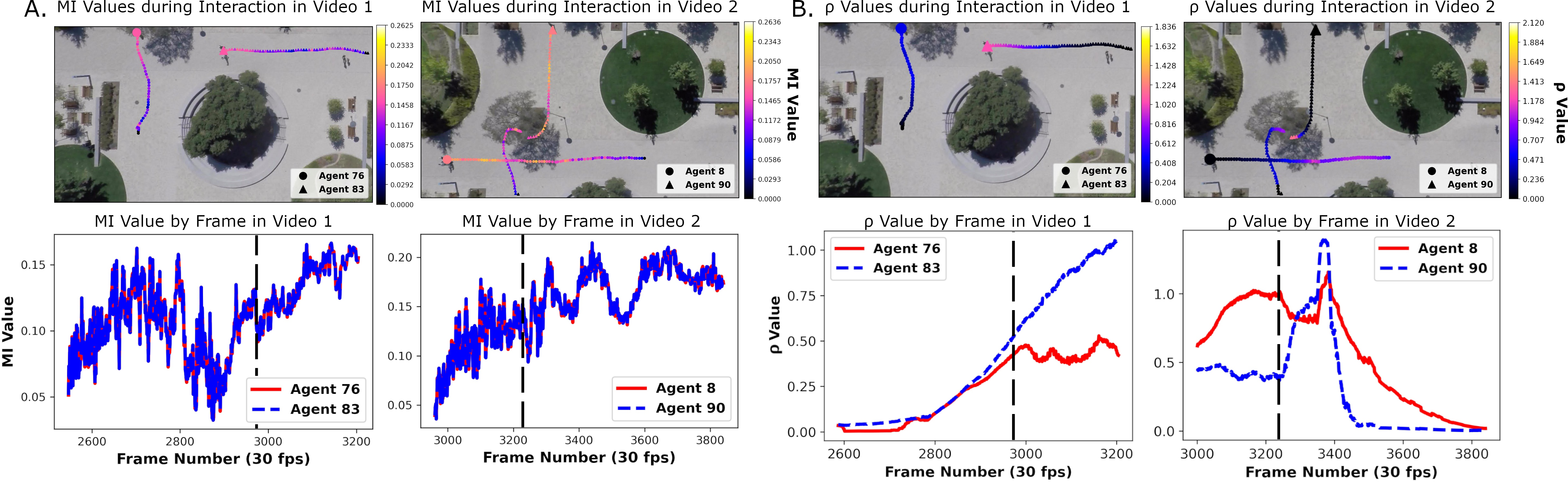}
\caption{The trajectories of the same two individuals are plotted from two videos in the SDD's Coupa scene. A portion of this interaction is included in both videos approximately between frames 3000 and 3200. MI takes different values over these frames for each video.
In contrast, the $\rho$ values which only rely on the most recent 30 frames at each time point show nearly identical values for each video between these overlapping frames. This demonstrates how the video overlaps may impact results differently depending on the methods being used.}
\label{fig.SceneBoundary}
\end{figure*}

When both time and location overlap, interactions that occur within the overlapping regions appear in multiple videos. This leads to redundancy in the observed interactions depending on the methods used. Fig.~\ref{fig.SceneBoundary} uses MI and $\rho$ to demonstrate how this overlap may affect different measures. The MI values within the overlapping regions differ between the two videos, while the values of $\rho$ are highly similar. 

The discrepancy between MI values is explained since MI relies on the full past of the trajectories, leading to higher MI values in video 1 during the overlapping frames as there is more past trajectory data providing information about the current interaction. Since $\rho$ only relies on the past one second of information, the interactions result in nearly identical values of $\rho$. This demonstrates how measures such as our $\rho$, which care only about local information, lead to redundant values in overlapping sections of different videos. 

Several videos in the SDD overlap both in location and recording time, which may influence how certain applications such as object tracking or trajectory prediction are used on the dataset (e.g. users may assess the accuracy and consistency of tracking a single individual across two videos, while trajectory prediction models may need to split their training and test data accordingly). 

When only the time of recording is the same, the effects are less predictable. If the scenes are nearby (for instance within a college campus), then behaviors associated with a given time would affect multiple recordings such as rush hour traffic or students going to class. In the inD dataset, all recordings of a given intersection record the same location at different times. Because of this, the features within that location are present in each of the recordings belonging to that scene. For features that significantly impact behavior, such as bus stops or benches, this redundancy could disproportionately emphasize those features. 

\begin{figure}[h!]
\centering
\includegraphics[scale=0.225]{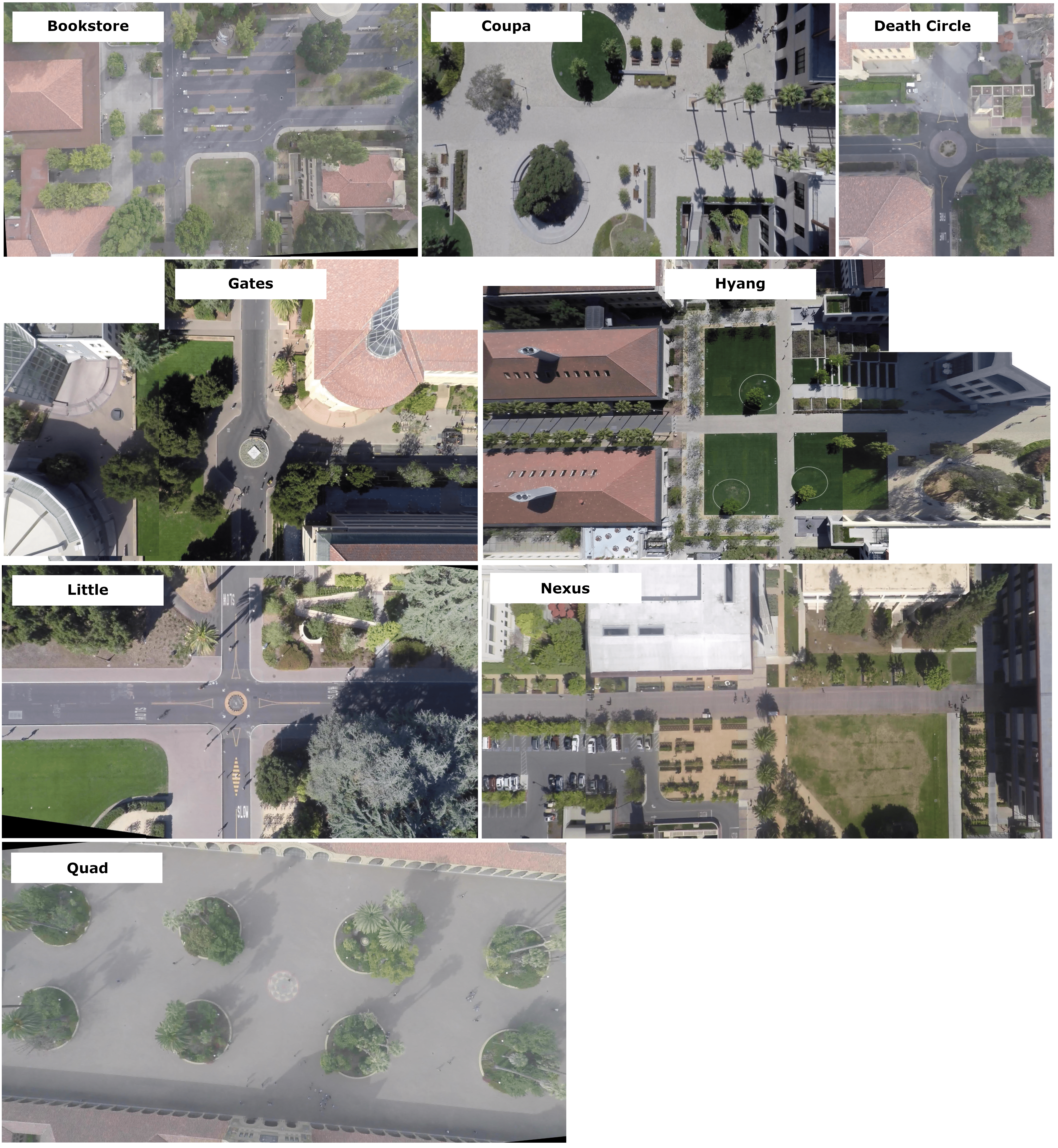}
\caption{Here we provide composite images of each scene as an aid for users of the SDD to more easily understand where each video is in relation to the others within a given scene.}
\label{composites}
\end{figure}

The groups of videos in each scene which overlap in time or location for each scene in the SDD are listed in Table \ref{table:overlaps}. In this table, "partial" overlap indicates that only a subset of the videos within the scene overlap either in time or location. We list which videos were shot simultaneously in the "Split Videos" data. The cause for such videos appears to be that certain videos are shot over a larger area prior to being broken into multiple videos, each covering a smaller subset of the area from the original recording. This is most clear with Coupa as seen in Fig. \ref{fig.CoupaOverlap}. We additionally provide the composite images of each scene in Fig. \ref{composites} for reference. These composite images allow users of the dataset to better visualize how individual videos relate to one another within a scene to identify where these characteristic overlaps occur.

\begin{table}
\caption{Outline of overlapping and split videos in the SDD.\\}
\centering
\scalebox{.7}
{
\vspace{1cm}
\begin{tabular}{@{}lcccccccc@{}}
\cline{1-9}
{\bf Overlap} & {\bf Bookstore} & {\bf Coupa} & {\bf DeathCircle} & {\bf Gates} & {\bf Hyang} & {\bf Little} & {\bf Nexus} & {\bf Quad}\\
\cline{1-9}
Location & Partial & Partial & Full & Partial & Partial & Partial & Partial & Partial\\ 
Time & Partial & Full & None & Partial & Partial & Partial & Partial & Full\\ 
\cline{1-9}
Split Videos & 1-6 & 1-4 & None & 0-2 & 6,10-14 & 1-3 & 0-2 & 0-3\\
 & & & & 4-7 & 7-9 & & 3-5 & \\
 & & & & 5,6 & 2,3 & & 6-8 &\\
 & & & & & & & 9-11 &\\
\cline{1-9}
\end{tabular}}
\label{table:overlaps}
\end{table}

\section{Discussion and Conclusion}

In this work, we have demonstrated several key characteristics of the SDD which have not been properly addressed before. These characteristics are: the properties of the lost annotations, differences in class diversity and behavior compared with the inD dataset, and the scene layout. We have provided clear and intuitive visualizations and descriptions of each, including the potential impacts they have in different applications. 

For those that have previously used the SDD without accounting for these characteristics, it is possible that the accuracy of their model was worsened by erroneous preprocessing or because the SDD was not well suited to their methods. In this paper, we have shown the potential for the former case by leaving the lost annotations in during preprocessing for the trajectory prediction models PECNet and Y-Net (Table \ref{pecnet}), which resulted in a worsened ability of the networks to generalize to the inD dataset when trained on the SDD compared to models trained without the lost annotations being included. We have also provided a comparison to the inD dataset to help future researchers make informed decisions regarding which data best fits their methods. We believe that these contributions will not only make it easier to work with the SDD, but also to do so properly while being able to better interpret the results obtained regardless of the application.

Understanding the properties of the SDD's annotated data is vital when using it, as detailing how your data is preprocessed (whether or not you have removed lost annotations, how you handle trajectories which become split as a result, etc), significantly improves others' ability to reproduce your results. Despite this, information surrounding these steps is frequently left out of publications. While the presence of split trajectories is worth noting and more difficult to identify, we suspect that it is also less impactful than the lost coordinates. Any model which is not expected to recognize that all four partial trajectories in Fig. \ref{fig.SplitTrajectory} belong to the same pedestrian (whose class changes from pedestrian to biker mid-trajectory) may train just as well by treating them as four unique individuals with non-overlapping frames. 

The differences in classes between the two datasets has implications for when each dataset is best suited for a given application. Agents navigating the SDD are less adherent to scene features such as sidewalks, and there are far fewer moving cars. This suggests that the SDD would be well suited for applications intended for use in areas such as parks or shopping centers, where pedestrians make up most of the traffic and are less observant of scene features. By contrast, the inD dataset may be better suited for street traffic where the prevalence of moving cars forces foot-traffic to more strictly adhere to sidewalks.

Lastly our demonstration of the overlap between videos in the SDD is useful for applications which can take advantage of this characteristic. An example of this is for tracking, where the overlapping locations could be used to test a method's ability to track an individual's trajectory across multiple adjacent videos. This information can be used by those looking to utilize different training/testing splits for the SDD data. Most trajectory prediction papers using the SDD utilize the Trajnet split outlined in \cite{sadeghian2018trajnet}. For the most part this split accounts for overlapping scene locations and simultaneous recordings, however we provide information in Table \ref{table:overlaps} regarding which scenes contain overlapping locations or are recorded simultaneously in case other users of the dataset want to investigate alternative splits. This information is supplemented by the composite images of each scene provided in Fig. \ref{composites}, which as best we could determine have yet to be published or provided anywhere. While simple, this information is not readily clear when using the SDD, and we've included it in keeping with this work's goal of improving the usability of the SDD.

\section*{Acknowledgment}
This work has been partially supported by NSF 2053480; the findings are those of the authors only and do not represent any position of these funding bodies.

\renewcommand{\refname}{References}
\bibliographystyle{IEEEbib}

\end{document}